\documentclass[10pt,twocolumn,letterpaper]{article}

\usepackage{cvpr}              

\usepackage{graphicx}
\usepackage{amsmath}
\usepackage{amssymb}
\usepackage{booktabs}
\usepackage[accsupp]{axessibility}

\usepackage[pagebackref,breaklinks,colorlinks]{hyperref}
\usepackage{tabularx}

\usepackage[capitalize]{cleveref}
\crefname{section}{Sec.}{Secs.}
\Crefname{section}{Section}{Sections}
\Crefname{table}{Table}{Tables}
\crefname{table}{Tab.}{Tabs.}

\def\cvprPaperID{*****} 
\def\confName{CVPR}
\def\confYear{2023}

\begin{document}

\title{FewSOME: One-Class Few Shot Anomaly Detection with Siamese Networks}

\author{Niamh Belton$^{1,2}$\\
\and
Misgina Tsighe Hagos$^{1,3}$\\
\and
Aonghus Lawlor$^{3,4}$\\
\and
Kathleen M. Curran$^{1,2}$\\ 
\and 
$^{1}$Science Foundation Ireland Centre for Research Training in Machine Learning \\
$^{2}$School of Medicine, University College Dublin \\ 
$^{3}$School of Computer Science, University College Dublin \\ 
$^{4}$Insight Centre for Data Analytics, University College Dublin, Dublin, Ireland\\
{\tt\small \{niamh.belton, misgina.hagos\}@ucdconnect.ie\ \{aonghus.lawlor, kathleen.curran\}@ucd.ie}
}

\maketitle

\begin{abstract}

Recent Anomaly Detection techniques have progressed the field considerably but at the cost of increasingly complex training pipelines. Such techniques require large amounts of training data, resulting in computationally expensive algorithms that are unsuitable for settings where only a small amount of normal samples are available for training. We propose \lq{}\textbf{Few} \textbf{S}hot an\textbf{OM}aly d\textbf{E}tection\rq{} (FewSOME), a deep One-Class Anomaly Detection algorithm with the ability to accurately detect anomalies having trained on \lq{}few\rq{} examples of the normal class and no examples of the anomalous class. We describe FewSOME to be of low complexity given its low data requirement and short training time. FewSOME is aided by pretrained weights with an architecture based on Siamese Networks. By means of an ablation study, we demonstrate how our proposed loss, \lq{}Stop Loss\rq{}, improves the robustness of FewSOME. Our experiments demonstrate that FewSOME performs at state-of-the-art level on benchmark datasets MNIST, CIFAR-10, F-MNIST and MVTec AD while training on only 30 normal samples, a minute fraction of the data that existing methods are trained on. Moreover, our experiments show FewSOME to be robust to contaminated datasets. We also report F1 score and balanced accuracy in addition to AUC as a benchmark for future techniques to be compared against. Code available; \url{https://github.com/niamhbelton/FewSOME}.

\end{abstract}

\section{Introduction}
\label{sec:intro}

Anomaly Detection (AD) refers to any technique that attempts to detect samples that are substantially distinct from the majority of other samples in a dataset. In recent years, Machine Learning algorithms for AD have become increasingly accurate, but at the cost of increased complexity. Self-supervised techniques, in particular, have made significant breakthroughs in terms of accuracy. However, most of these methods have complicated training pipelines that involve performing extensive transformations and augmentations to the data during training. Such complex algorithms are not suitable for the specific few-shot setting where only a few examples (i.e. shots) of the normal class exist and no examples of the anomalous class exist. In this paper, we propose \lq{}\textbf{Few} \textbf{S}hot an\textbf{OM}aly d\textbf{E}tection\rq{} (FewSOME), a deep one-class AD algorithm that performs at State-of-the-Art (SOTA) level with a fraction of the complexity of existing methods in terms of training data size and training time. FewSOME has a Siamese-like architecture that consists of multiple branches of neural networks where each neural network has shared weights. Data samples are input in tandem into each neural network during training with the objective of transforming the data samples to representations that are within close proximity of eachother. This architecture is suitable for AD given that Siamese Networks are known for making predictions about unknown class distributions \cite{siamese}. As is common in recent AD techniques, FewSOME benefits from weight initialisation on the ImageNet dataset \cite{imagenet,igd,panda}. FewSOME is trained using our proposed loss, \lq{}Stop Loss\rq{} denoted as $L_{stop}$ to prevent \lq{}Representational Collapse\rq{}. Representational Collapse is a known issue in one-class AD tasks where the model learns to map all inputs to a constant output. Our proposed loss, \lq{}Stop Loss\rq{}, can prevent this and improve the robustness of the model.

In this paper, we present FewSOME, a new SOTA in the field of Few Shot Anomaly Detection (FSAD) on normal data samples. In sections \ref{compar} and \ref{small_data}, we show that FewSOME can perform competitively with SOTA techniques that were trained in the classical AD setting (i.e. trained on large datasets) whilst having significantly lower complexity in terms of training data size and training time. Our experiments show that FewSOME is robust to contaminated datasets (section \ref{contam}) and by means of an ablation study, we show how our proposed loss \lq{}Stop Loss\rq{} boosts model performance and prevents representational collapse (section \ref{alpha}). Additionally, we report F1 score and balanced accuracy in addition to Area Under the Curve (AUC) as a benchmark for future techniques to be compared against.





\section{Related Work}
\label{rel}

Traditionally, AD was performed using simple Machine Learning approaches such as Isolation Forest \cite{if} or statistical methods such as Kernel Density Estimation (KDE) \cite{kde} and One-Class Support Vector Machine (OC-SVM) \cite{ocsvm}. However, the requirement for manual feature engineering and the poor computational scalability of these approaches led to the exploration of more advanced techniques for AD.

\textbf{Deep One-Class}. Deep One-Class methods aim to learn the features of normality in a dataset and classify data samples that diverge substantially from the extracted normal features as anomalies. DeepSVDD \cite{deepsvdd} is one of the most commonly known one-class AD techniques. It trains a network that transforms the training samples to a representation space where normal samples are contained inside a hypersphere. Data samples whose representations fall outside the radius of the hypersphere are considered to be anomalies. The Deep Robust One-Class Classifcation (DROCC) \cite{drocc} significantly outperformed DeepSVDD by margins up to 20\% on some AD tasks by generating synthetic anomalies on which the model could train on. A more recent method, PANDA \cite{panda}, shows that SOTA AD performance can be achieved by training large neural network architectures of ResNet-152 and WideResNet-50 with pretrained ImageNet weights  \cite{imagenet}.  The newly proposed, Interpolated Gaussian Descriptor (IGD) \cite{igd} trains a Gaussian classifier, initialised with pretrained weights on the ImageNet dataset, to minimise the distance between representations of normal images and the centre of the normal image distribution whilst also adversarially interpolating training samples. They implement a critic module to constrain the training of IGD to be based on normal samples that are representative of the majority of normal samples, rather than anomalous samples that may be present in the training data.

\textbf{Generative}. Deep Autoencoders and Generative Adversarial Networks (GANs) have dominated the field of AD in previous years \cite{dcae,minlgan,memae,ocgan,lsa,anogan,cavga}. Autoencoders learn a mapping from the input to a latent space of lower dimensionality. The images are then reconstructed from the latent space and anomalies are identified by large reconstruction errors. GANs for AD learn to generate normal samples. During testing, the model attempts to match the test sample to a point in the generator's latent space. It then reconstructs the test sample based on this point and anomalies are identified by large reconstruction errors. Although Autoencoders and GANs can accurately detect anomalies, they have various limitations; it can be difficult to estimate the intrinsic dimensionality of the data i.e. it requires manual selection of the latent space dimensionality, they do not specifically target anomaly detection, they often have a significant number of network parameters, they require large amounts of training data and it is computationally expensive to reconstruct the entire image.


\textbf{Self-Supervised.} Self-Supervised techniques have progressed the AD field considerably in terms of accuracy. GEOM \cite{geom}, GOAD \cite{goad} and RotNet \cite{rotnet} are examples of transformation based methods that apply transforms to the training data such as flipping and rotating. They then train a classifier to predict the transform applied. The training sample is classified as normal if the model can predict the transform and an anomaly otherwise. Another method, Contrasting Shifted Instances (CSI) \cite{csi}, contrasts each training sample with an augmentation of itself, where the augmentation is considered to be a different class to the original training sample. CSI can therefore, constrastively learn meaningful representations of the normal class. Although these techniques significantly outperform existing methods, recent literature under-reports their limitations such as the requirement for domain specific transformations. For example, augmentations such as flipping an image of class \lq{}6\rq{} will convert this to class \lq{}9\rq{} in the handwritten digit dataset, MNIST. Therefore, many of the recent approaches have not tested their technique on the standard AD benchmark MNIST dataset, highlighting their lack of robustness. It was also previously shown that the performance of GEOM significantly decreases if applied to data that has been augmented, whilst the performance of the one-class method DROCC was unaffected \cite{drocc}. These transformations/augmentations also increase the complexity of the training and testing pipeline with CSI requiring up to 40 augmentations at test time and thus, limiting their application in the real-world.

\textbf{Few-Shot Anomaly Detection (FSAD).} DeepSAD \cite{deepsad}, an extension of DeepSVDD, was one of the first AD techniques to adapt their loss function and training strategy to consider labelled normal and anomalous data. Later studies, focused on the FSAD setting where there is a large amount of unlabelled samples and few shots of the anomalous class available for training \cite{devnet,colon,siam_att,zero_exp}. Liznerski et al. (2022) \cite{zero_exp} studied the zero-shot setting where the model trained on all available unlabelled data but no anomalies. They use pretrained weights from the Contrastive Language-Image Pre-training (CLIP) \cite{clip} model. CLIP was trained on a large dataset of image and text pairs with the objective of related pairs being within close proximity to eachother in representation space and unrelated pairs being distant from eachother in representation space. Other works have studied the separate FSAD setting where multiple normal classes exist in the data and there is only a few shots of each normal class available for training \cite{registration,normality}. These techniques aim to not only identify anomalies that are distant from the normal data but they also aim to identify anomalies that exist in between the normal classes. FewSOME tackles the less studied FSAD setting, that requires no level of supervision, of training on only few examples of the normal class. Recent works have studied this setting for the specific use case of detecting anomalies in industrial settings \cite{texture,MAEDAY}. They also study the zero-shot setting, where the model has not seen any examples of the normal or anomalous class but detect anomalies by identifying image regions that break the homogeneity of the input data sample. The first work to train on few shots of the normal class for more general AD was the Hierarchical Transformation-Discriminating Generative (HTDG) model \cite{hier_few}. They employ a generative model and use self-supervision to perform AD based on only one, five and ten shots of the normal data. Unlike FewSOME, they compare their performance to competing methods trained only in the few shot setting, whilst we show in sections \ref{compar} and \ref{small_data} that FewSOME's performance surpasses that of HTDG and FewSOME also performs at the same level of existing SOTA that were trained in the classical AD setting (i.e. trained on large datasets).

\textbf{Representational Collapse.} Techniques with examples of the anomalous class or techniques that synthetically generate anomalous samples can learn meaningful representations of the normal class through contrastive learning. However, one-class AD algorithms can be susceptible to Representational Collapse as their objective to minimise the representation space between training samples can be easily satisfied by learning to map all inputs to a constant output. One-Class method, DeepSVDD prevents Representational Collapse by removing bias terms and freezing the value of its hypersphere centre during training. PANDA prevents collapse by using early stopping and another technique known as \lq{}elastic regularisation\rq{}. Relatedly in the field of visual representation learning, a recent technique known as \lq{}Stop Gradient\rq{} has been empirically proven to avoid Representational Collapse by preventing the backflow of gradients through one of the branches of a Siamese network \cite{simsiam}.

\section{FewSOME}

\begin{figure*}[t]
  \includegraphics[width=\textwidth]{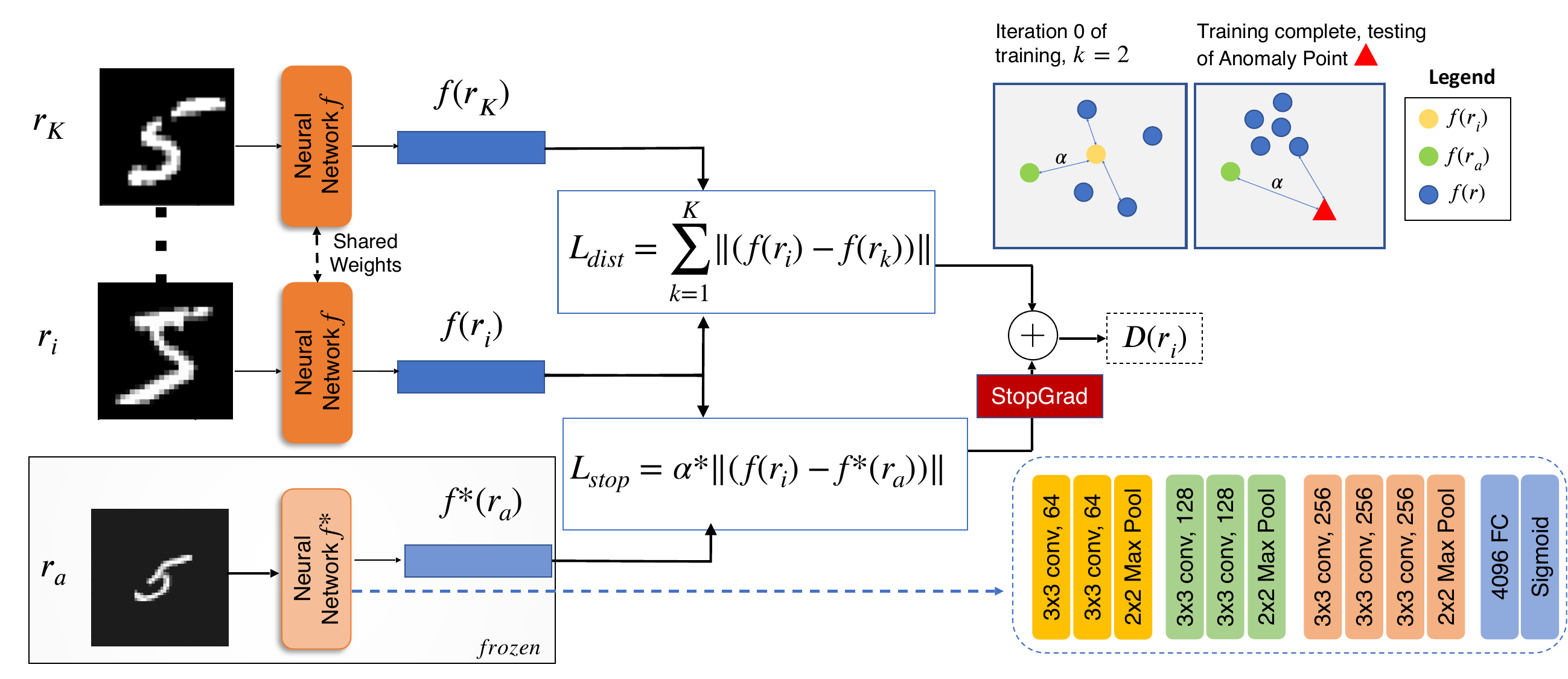}
  \caption{Training process for the MNIST dataset when the normal class is equal to five. In each iteration, $r_i \in\ R_N$ and $\{r_1,...r_K\} \in\ R_N$ are input into the network $f$ to obtain the feature embeddings, $f(r_i)$ and $\{f(r_1),...f(r_K)\}$. The values of $f^{*}(r_a)$ are obtained in the first initial pass through the model and they are unchanged throughout training. The $L_{dist}$ is calculated in parallel with $L_{stop}$ to obtain $D(r_i)$. The model is trained with the objective of minimising $D(r_i)$. The VGG-3 architecture is shown in the figure where \lq{}conv\rq{} refers to a convolution, \lq{}Max Pool\rq{} refers to a Max Pooling operation and \lq{}FC\rq{} refers to a Fully Connected layer. The figure also depicts a hypothetical example of the feature embeddings projected into 2D space. The blue dots are the feature embeddings of the Reference Set. The green dot is $r_a \in\ R_N$, the yellow dot is $r_i \in\ R_N$ and the red triangle is an anomaly during testing. Before training all feature embeddings are dispersed in the 2D space. After training, the feature embeddings of the Reference Set are closer together, while the anomalous sample is distant from the Reference Set.}

  \label{sense}
\end{figure*}

This section outlines FewSOME, visualised in Figure \ref{sense}. Given a dataset $X \subseteq \mathbb{R}^{d}$, a sample of size $N$ is randomly sampled from $X$ so that the training data is $R_N = \{r_1,....,r_N\} \subseteq X$. This is referred to as the \lq{}Reference Set\rq{}. This is different to the standard training set as the size of the Reference Set is typically a small fraction of the size of the training set. A neural network, $f$, is then trained to transform the input space $R_{N} \subseteq \mathbb{R}^{d}$ to the representation space $f(R_{N}) \subseteq \mathbb{R}^{l}$. The representation space of $R_N$ is denoted as $f(R_N) = \{f(r_1),....,f(r_N)\}$, where $f(r_i)$ is the representation of $r_i$  in the form of a 1D feature embedding with dimensions $1 \times l$. The objective of the network, $f$, is to learn weights $W$ that minimise the Euclidean Distance (ED) between feature embeddings of the Reference Set. This is achieved by employing a distance-based loss function. The distance component of the loss function, $D(r)$ consists of $L_{dist}$ and Stop Loss, denoted as $L_{stop}$ (equation \ref{eq1}). In each epoch of the training process, $D(r_i)$ is computed for each $r_i \in R_N$.

\begin{equation}
\label{eq1}
\small
 D(r_{i}) = \underbrace{\sum_{k=1}^{K}\Big(||f(r_i) - f(r_{k})||\Big)}_{L_{dist}} + \underbrace{\alpha ||f(r_i) - stop(f^{*}(r_{a}))||}_{L_{stop}}
\end{equation}

\subsection{\textbf{\boldmath$L_{dist}$} } The $L_{dist}$ component represents the Siamese architecture of FewSOME. FewSOME differs from the typical Siamese Network as it trains on only one class and the number of branches in the network is a hyper-parameter of the model. 
The hyper-parameter $K$ is the number of data samples in $R_N$ that are input into $f$ in tandem with $r_i$. The term $||f(.) - f(.)||$ is the ED between two feature embeddings divided by $\sqrt{l} $ where $l$ is the dimension of the 1D feature embeddings. As the final layer of $f$ is a Sigmoid layer, it can be proven that dividing the ED by $\sqrt{l}$ results in the term $ ||f(.) - f(.)||$ being bounded between zero and one.

For each $r_i$ during training, $r_{k}$ 
is selected randomly. 
A variation of FewSOME, named Smart FewSOME (S-FewSOME) selects $r_k$ such that $r_{k} = argmax_{r_{k}}||f(r_{i})-f(r_k)||$ where $r_i$ and $r_k$ $\in R_N$. This is commonly done in distance based loss functions such as triplet loss \cite{pairs}. This can speed up the convergence of the model, however, it increases the complexity of the model. 

\subsection{\textbf{\boldmath$L_{stop}$} } 

The $L_{stop}$ component of $D(r_i)$ computes the distance between $f(r_i)$ and $f(r_a)$ where $r_a \in R_N$. The data sample, $r_a$ is used to compute the \lq{}anchor\rq{}, $f(r_a)$. It is arbitrarily selected from $R_N$ prior to training. The values of $f(r_a)$ are computed as the first initial pass through the model. Stop Gradient \cite{simsiam}, denoted as $stop$ in equation \ref{eq1} is then applied to $f(r_a)$. This prevents the back-flow of gradients i.e. the weights are updated so that $f(r_i)$ is moved closer to $f(r_a)$, rather than the weights updating to move both $f(r_i)$ and $f(r_a)$ closer to each other. This results in the value of $f(r_a)$ being frozen for the duration of model training. Frozen values are denoted by * in $f^{*}(r_a)$. We found that the selection of $r_a$ had an insignificant impact on model performance and model convergence (see also section \ref{sensitivity}). Hyper-parameter $\alpha$, $0\leq \alpha \leq 1$, controls the trade-off between $L_{dist}$ and $L_{stop}$. The anchor in FewSOME behaves differently to DeepSVDD's frozen centre. For example, if $\alpha < 1$, minimising the distance from $f(r_i)$ to the anchor $f(r_a)$ is not the primary objective when minimising $D(r_i)$ and therefore, the anchor does not necessarily become the centre. Our proposed loss is also less complex as the anchor is computed based on a randomly selected training sample and it does not require the representations for all training samples to be calculated in order to assign its value, as is required when assigning the value of DeepSVDD's centre, $c$.


Equation \ref{eq2} shows the objective function. The first term is equal to Contrastive loss\footnote[1]{$Contrastive\ Loss = \frac{1}{N}\sum_{i=1}^{N}\frac{1}{2} (1-y_i) (||(x_{i}-x_{j}||)^{2} + \frac{1}{2} (y_i) MAX(0, margin - (||(x_{i}-x_{j}||)^{2}$, where $N$ is the number of samples, $y$ is the label ($y=0$ when both samples are the same, $y=1$ when two samples are different), $||(x_{i}-x_{j}||$ is the ED between two data samples $x_{i}$ and $x_j$ and the margin is an arbitrary small number.} when $y=0$. During training, $y$ will always equal zero in FewSOME as all samples in $R_N$ are of the same class. The last term shows the weight decay regularization where $0<= \lambda <= 1$.  

\begin{equation}
\label{eq2}
\min_{D(r_i), W} \frac{1}{N}\sum_{i=1}^{N}\frac{1}{2}D(r_{i})^{2} +  \frac{\lambda}{2}||W||^{2} 
\end{equation}

During testing, samples from a test set $T \subseteq \mathbb{R}^{d}$ are input into $f$ to transform them to the representation space. The transformed test samples are then ranked by an anomaly scoring function, $s$. 
Existing anomaly scoring functions score a test sample based on its distance to the centre of the normal class representation space. 
We found that assigning an anomaly score to test point, $t_i \in T$, based on the distance between $f(t_i)$ and the nearest representation in the Reference Set, $f(r) \in f(R_N)$, and the distance from $f(t_i)$ to the anchor, $f^*(r_a)$ gave optimal results (equation \ref{eq4}).

\begin{equation}
\label{eq4}
s(t_{i}) = argmin_{f(r)}\{||f(t_{i}) - f(r)|| + \alpha||f(t_i) - f^{*}(r_a)||\}
\end{equation}

\subsection{Properties to Prevent Representational Collapse}
\label{rep_col}

We define Representational Collapse as the case in which all feature embeddings of the Reference Set are equal to the mean feature embedding of the Reference Set i.e. $\forall \ r \in R_N: \frac{1}{N}\sum_{i=1}^{N}f({r_i}) = f(r)$. 
In simpler terms, collapse results in the model mapping all inputs to a constant output. We outline two scenarios in which collapse could occur and the measures that were taken to prevent it; 

\begin{enumerate}
\item \textbf{In the absence of $\mathbf{L_{stop}}$:} if the weights of any layer in $f$ converge to all zeros, this would result in $\forall \ r \in R_N: \frac{1}{N}\sum_{i=1}^{N}f({r_i}) = f(r)$. Although the model has not learned the features of normality, it has in theory achieved its objective of minimising the representation space with $L_{dist}=0$. However, we prevent this scenario by including $L_{stop}$ which minimises the distance between $f(R_N)$ and non-zero value $f^{*}(r_a)$. This ensures that the model has not achieved its objective by outputting constant representations.
\item \textbf{In the presence of $\mathbf {L_{stop}}$:} if the weights of any layer in $f$ converge to all zeros and bias terms exist, $f$ can achieve its objective by updating the bias terms so that each feature embedding is equal to the feature embedding of the anchor i.e. $\frac{1}{N}\sum_{i=1}^{N}f({r_i}) = f^{*}(r_a)$  . Similar to DeepSVDD, we prevent this type of representational collapse by removing bias terms. 
\end{enumerate}

\section{Experiments}

FewSOME was evaluated on four benchmark datasets, MNIST \cite{mnist}, Fashion-MNIST (F-MNIST) \cite{fmnist}, CIFAR-10 \cite{cifar} and MVTec AD \cite{mvtec}. The former three datasets consist of ten classes. Following standard AD protocol, each class was set to normal and all other classes were set to anomalies. MVTec AD has 15 classes consisting of high-resolution images of different industry objects and textures. Within each class, there is a training set of normal images and a test set of normal and anomalous samples. Ten models were trained for each normal class. Each of the ten models were trained on different randomly sampled Reference sets from the provided training data.

\subsection{FewSOME Implementation Details}
The model backbone for MNIST, CIFAR-10 and F-MNIST is a simple architecture we name \lq{}VGG-3\rq{} based on the first three blocks of VGG-16. Following the training protocol of previous SOTA DROCC, 20\% of the official training data was used as a validation set to tune the model parameters. The hyper-parameters of FewSOME include the Reference Set size, $N \in \{30, 60\}$, $K \in \{1 , 2, 3\}$, feature embedding dimension, $l \in \{1024, 2048\}$, batch size $\in \{1 , 8, 16, 30\}$,  $\alpha$ in the range $ [0.01, 1]$ and learning rate in the range $[1e-2, 1e-8]$. The early stopping criteria for FewSOME is to stop training once the rate at which the training loss is decreasing is less than 0.5\% for a patience of two. The optimiser was Adam \cite{adam}. The image dimensions in MVTec AD range from $700\times700$ to $1024\times1024$. These images were rescaled to $128\times128$. A larger architecture, ResNet-18 \cite{resnet} was then employed as the model backbone and the feature embedding dimension, $l$ was set to $1000$. For MVTec, the average best result (averaged over ten seeds) on the test set was reported as was done for competing methods for fair comparison. Weights were initialised using Kaiming initialisation or with weights pretrained on ImageNet.

\subsection{Competing Methods Implementation Details}
We compare FewSOME against typical baseline methods. We primarily compare FewSOME against DeepSVDD and DROCC as they are proven SOTA also belonging to the \lq{}Deep One-Class\rq{} class of AD techniques. As the original papers do not report on all benchmark datasets, we implement both models for F-MNIST and MVTec and we additionally implement DROCC for MNIST. We tuned DROCC's hyper-parameters using grid-search on a validation set of 20\% of the provided training data for MNIST, CIFAR-10 and F-MNIST. For MVTec, the average best result (averaged over ten seeds) on the test set was reported. For MVTec, DROCC's model backbone was scaled up from a LeNet architecture to a ResNet-18. As DeepSVDD requires pretraining of an autoeconder, it was too computationally expensive to increase the size of its architecture to a ResNet-18 for training on MVTec. Therefore, the images were further downsized to $64\times64$ and we increased the number of model parameters by two million by adding convolutional layers and kernels. Although newer techniques such as PANDA and IGD also belong to the \lq{}Deep One-Class\rq{} category, it would not be a fair comparison to compare them to FewSOME given FewSOME's simple training pipeline and smaller architecture. The recent technique, PANDA employs ResNet-152 and WideResNet-50 architectures. IGD additionally has a complex training pipeline that involves reconstructing the images and adversarially generating training samples.

\section{Results}

\label{results}

\subsection{Comparative Analysis}
\label{compar}

Table \ref{tab:res} shows the Area Under the Curve (AUC) averaged over all classes (ten seeds per class) compared with other SOTA on the provided test sets. Figures with an asterick are reported based on our implementation. FewSOME and the Smart FewSOME (S-FewSOME) variation show substantial performance improvements over baseline and generative models with training data sizes of $N=30$ for MNIST, CIFAR-10 and F-MNIST and $N=60$ for MVTec, a fraction of the training data that competing methods use (competing methods train on 5,000 images for CIFAR-10 and 6,000 images for MNIST and F-MNIST). It can be noted that S-FewSOME results in only marginal increases in performance in some cases compared to standard FewSOME, indicating that the lower complexity standard FewSOME model is sufficient. As it is known that 90\% of the test data are anomalies, test samples with anomaly scores in the 10th percentile and above are classified as anomalies for all models. The F1 score and balanced accuracy were then calculated, averaged across all classes and reported in Table \ref{tab:f1}. Although FewSOME and DROCC perform similarly for CIFAR-10 in terms of AUC, Table \ref{tab:f1} demonstrates that FewSOME outperforms it in terms of F1 score and balanced accuracy. This highlights the necessity for metrics in addition to AUC to be reported with all AD algorithms. 

\begin{table}[t] 
\small
\centering
  \caption{AUCs in \% averaged over each normal class (ten seeds per class) on MNIST, CIFAR-10, F-MNIST and MVTec AD. * Results based on our implementation. $^{\dagger}$ Few-Shot learning approaches (trained on few shots of normal data).}
  \label{tab:res}
  \begin{tabular}{c|c|c|c|cccc}
    \hline
     Model & MNIST & CIFAR-10 & F-MNIST & MVTec\\
    \hline
    \hline
    \textbf{Baselines} \\
    OC-SVM\cite{ocsvm} &91.3& 64.8 &$\times$& $\times$

\\
    KDE \cite{kde} &86.9& 64.9 &$\times$& $\times$
    \\
    
    \textbf{Generative }&&&&\\ 
    DCAE \cite{dcae} &89.7& 59.4 &$\times$& $\times$
        \\
        MinLGAN\cite{minlgan} &$\times$& 68.7 &$\times$& $\times$
       \\
    LSA \cite{lsa}  &$\times$& 64.1 &$\times$&$\times$  \\
    
    OCGAN \cite{ocgan} &97.5& 65.7 &$\times$& $\times$
    \\
    AnoGAN \cite{anogan} &91.3& 61.8 &-&50.3 
    \\
    MEMAE \cite{memae} &97.5& 60.9 &$\times$&$\times$
    \\
    CAVGA \cite{cavga} &98.6& 73.7 &88.5& $\times$
 \\

    HTDG$^{\dagger}$ \cite{hier_few} &87.2& 70.2 &91.2& 78.0
 \\

    \textbf{One-Class }&&&&\\
    
    DeepSVDD \cite{deepsvdd} &92.4&  64.8 & $90.7$* & $72.2$*
    
    \\
    DROCC \cite{drocc} &87.8*&\textbf{76.9} & 90.5* & 74.5* 
  
    \\
    FewSOME$^{\dagger}$  &98.0& 76.6&\textbf{93.1} & \textbf{84.7}
    \\
    S-FewSOME$^{\dagger}$  &\textbf{98.1}& \textbf{76.9}&\textbf{93.1} & 82.3

   \\
   \hline

\end{tabular}

\end{table}
\

\subsection{Training on Few Normal Samples}
\label{small_data}
Five models were trained for each normal class in MNIST and CIFAR-10 with varying number of shots (i.e. training dataset sizes) ranging from two to 50. We repeat the experiments with competing methods DeepSVDD and DROCC. We also report the results of the previous SOTA in FSAD on normal data samples, HTDG. HTDG, as outlined in section \ref{rel}, is a generative model that uses self-supervision to detect anomalies. In this section, we introduce an additional competing method, IGD \cite{igd}. IGD, as outlined previously, is a deep one-class method that uses ImageNet weights and adversarially generates samples during training. They then classify anomalies by measuring their distance to a Gaussian centre of the normal class representation space. We include this as a competing method as they specifically demonstrate the performance of their method on small datasets.

\begin{table}[t]
\small
\centering
  \caption{F1 score and Balanced Accuracy in \% averaged over each normal class (ten seeds per class) on MNIST, CIFAR-10, F-MNIST and MVTec AD. All results are based on our implementation. The model \lq{}S-FewSOME\rq{} is the Smart FewSOME variation.}
  \label{tab:f1}
  \begin{tabular}{c|cccc}
    \hline

    &MNIST & CIFAR-10 & FMNIST & MVTec 
    \\
\hline
     \hline
    & \multicolumn{4}{c}{\textbf{F1}} \\

   DeepSVDD  &96.6&91.0& 95.4 &83.1
    \\
   DROCC &95.0& 92.1 &95.3&83.7
    \\
  FewSOME & \textbf{98.2} & 92.1 &\textbf{96.0}&\textbf{85.7}
    \\
  S-FewSOME & \textbf{98.2 }& \textbf{92.9 } &\textbf{96.0}&84.9
    \\
    \hline

    & \multicolumn{4}{c}{\textbf{Balanced Accuracy}}\\

       DeepSVDD &82.7&55.1& 76.8&57.1
    \\
   DROCC &73.7&58.0&76.2&58.0
    \\
  FewSOME & 90.9 & 60.6 &\textbf{80.2} & \textbf{63.2}
    \\
  S-FewSOME & \textbf{91.0} & \textbf{64.5} &\textbf{80.2} &61.6
    \\

\end{tabular}

\end{table}

For this experiment, FewSOME and DeepSVDD were trained until the rate at which the training loss was decreasing was less than 0.5\% for a patience of two. As DROCC does not benefit from pretrained weights, we used the same stopping criteria but with increased patience from two to ten. Due to IGD's architecture of a generator and critic module that behaves similar to a GAN discriminator, monitoring the training loss was an ineffective method for early stopping. Therefore, we evaluate it on the test set every 20 epochs and report the best result. Table \ref{tab:small_data} displays the AUC averaged over all normal classes for the varying training data sizes. There are three primary findings from Table \ref{tab:small_data}. Firstly, FewSOME achieves almost optimal performance on as little as $N=2$. Secondly, FewSOME outperforms the current SOTA in FSAD on normal examples, HTDG. Thirdly, FewSOME outperforms IGD, the SOTA for AD on small datasets. This is a notable result as IGD has a complex training pipeline that consists of a classifier, generator and a critic module that performs similar to a GAN discriminator, whilst FewSOME is a single network with a small number of model parameters ranging from four million to 11 million depending on whether VGG-3 or ResNet-18 is employed. 

\begin{figure*}[t]

  \includegraphics[width=.99\textwidth]{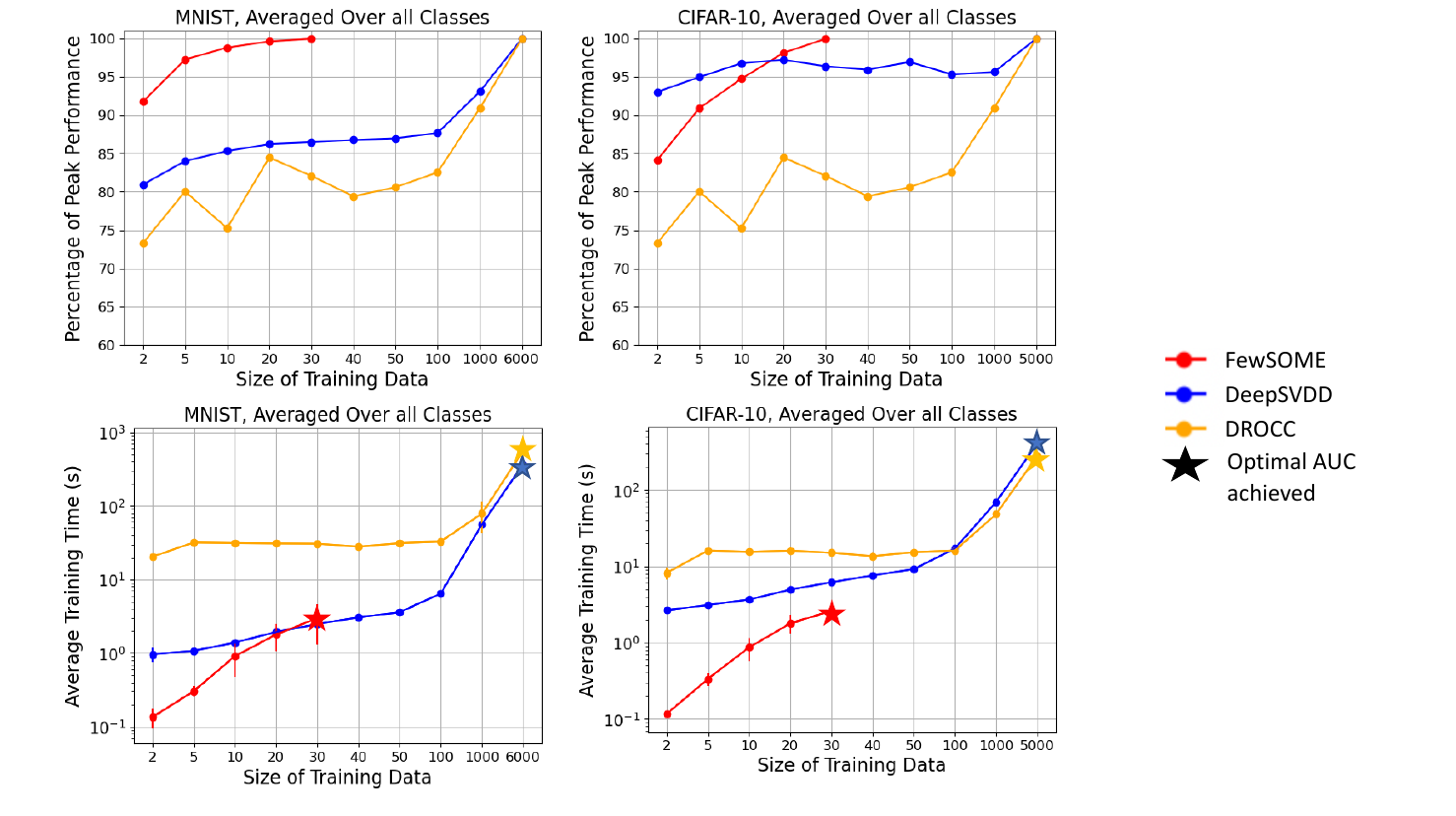}
  \caption{Top row: Percentage of peak performance achieved for each model at different training dataset sizes (averaged over all classes, 5 seeds per class) for MNIST and CIFAR-10.  Bottom row: The average training time for each model at different training dataset sizes (averaged over all classes, 5 seeds per class) for MNIST and CIFAR-10. Stars represent the training data size where the model achieves its peak AUC.}
\label{fig:small_data}
\end{figure*}

Figure \ref{fig:small_data} visualises how FewSOME reaches its peak performance when $N=30$, whilst both DROCC and DeepSVDD require all available training data to achieve their peak performance. The figure also shows FewSOME requires less training time than competing methods.

\begin{table}[t]
\small
\centering
  \caption{AUC in \% averaged over each normal class on MNIST and CIFAR-10 for shots (i.e. training set sizes) ranging from two to 50. All results are based on our implementation with the exception of HTDG. DeepSVDD has been abbreviated to DSVDD and FewSOME has been abbreviated to FSOME.}
  \label{tab:small_data}
  \begin{tabular}{c  |ccccccc}
    \hline
 Model & 2&5&10&20&30&40&50 \\
    \hline 
    \hline

    &\multicolumn{7} {c}{\textbf{MNIST}} 
    \\
        HTDG & $\times$& 85.9 & 87.2 & $\times$ & $\times$ & $\times$ & $\times$ \\

 DSVDD & $75.9$ & $78.8$ & 80.0 & 80.9 & 81.1 & 81.4 & 81.6
\\

DROCC   & 64.3& 70.3 & 66.1& 74.2 & 72.1 & 69.7 & 70.9 
\\
 IGD & 80.1 &83.4&88.5&90.2&92.8 &93.9 & 94.7 
\\
FSOME & \textbf{90.1} & \textbf{95.5} & \textbf{97.0} & \textbf{97.8} & \textbf{98.1} & \textbf{98.2} & \textbf{98.2} 
\\
\hline
    &\multicolumn{7} {c}{\textbf{CIFAR-10}} 
\\

    HTDG & $\times$& 67.5 & 70.2 & $\times$ & $\times$ & $\times$ & $\times$ \\

 DSVDD  & 57.4 & 58.6 & 59.7 & 60.0 & 59.4 & 59.2 & 59.8
\\

DROCC  & $54.2$ & $55.3$ & $55.6$ & $55.5$ & $56.2$   & $55.1$ & $55.5$ 
\\
IGD  &54.2 &58.2  &65.4&73.3&74.6 & 75.8 & 74.6
\\
FSOME  & \textbf{64.4} & \textbf{69.6} & \textbf{72.5} & \textbf{75.1} & \textbf{76.6} & \textbf{76.2} & \textbf{75.8} 
\\

\hline

\end{tabular}

\end{table}

\subsection{Contaminated Datasets}
\label{contam}
The following experiment was conducted by contaminating the training data with anomalies at rates of 1\%, 5\%, 10\% and 20\%. The results were averaged over normal classes. The results reported in Table \ref{tab:unsuper} show that FewSOME is robust to contamination in the data, achieving 96\% of its peak performance on both CIFAR-10 and MNIST despite 20\% of the Reference Set being contaminated with anomalies. At all levels of contamination, FewSOME is outperforming DeepSVDD and DROCC.

\begin{table} 

  \caption{AUC in \% averaged over each class (five seeds per class) at varying levels of contamination in the training data. Results based on our implementation.}
  \label{tab:unsuper}
  \setlength\tabcolsep{0pt}
\begin{tabular*}{\linewidth}{@{\extracolsep{\fill}} cc|cccc }
  \hline
   \multicolumn{2}{c|}{Model} &1\% & 5\% & 10\% &20\%  \\
  
    \hline
    \hline
    &&\multicolumn{4} {c}{\textbf{MNIST}} \\ 
 \multicolumn{2}{c|}{ DeepSVDD}  & 92.9 & 90.4 & 87.4 & 82.9 
    \\
    \multicolumn{2}{c|}{ DROCC} & 82.8 & 82.7 & 77.9 & 76.9 
  
    \\
      \multicolumn{2}{c|}{ FewSOME} &\textbf{97.2}&\textbf{96.7}&\textbf{95.8}&\textbf{93.7}
    \\
    \hline
  &  &\multicolumn{4} {c}{\textbf{CIFAR-10}} \\ 

     \multicolumn{2}{c|}{ DeepSVDD}  & 63.1 & 62.3 & 61.5 & 60.3
    \\
    \multicolumn{2}{c|}{ DROCC}  & 70.1 & 69.6 & 68.9 & 68.4
  
    \\
     \multicolumn{2}{c|}{ FewSOME} &\textbf{76.0} & \textbf{75.1} & \textbf{74.8} & \textbf{73.7}
    \\

\end{tabular*}
\end{table}

\section{Further Analysis}

\subsection{Ablation Study on $L_{stop}$}
\label{alpha}
\begin{figure}

  \includegraphics[width=\linewidth]{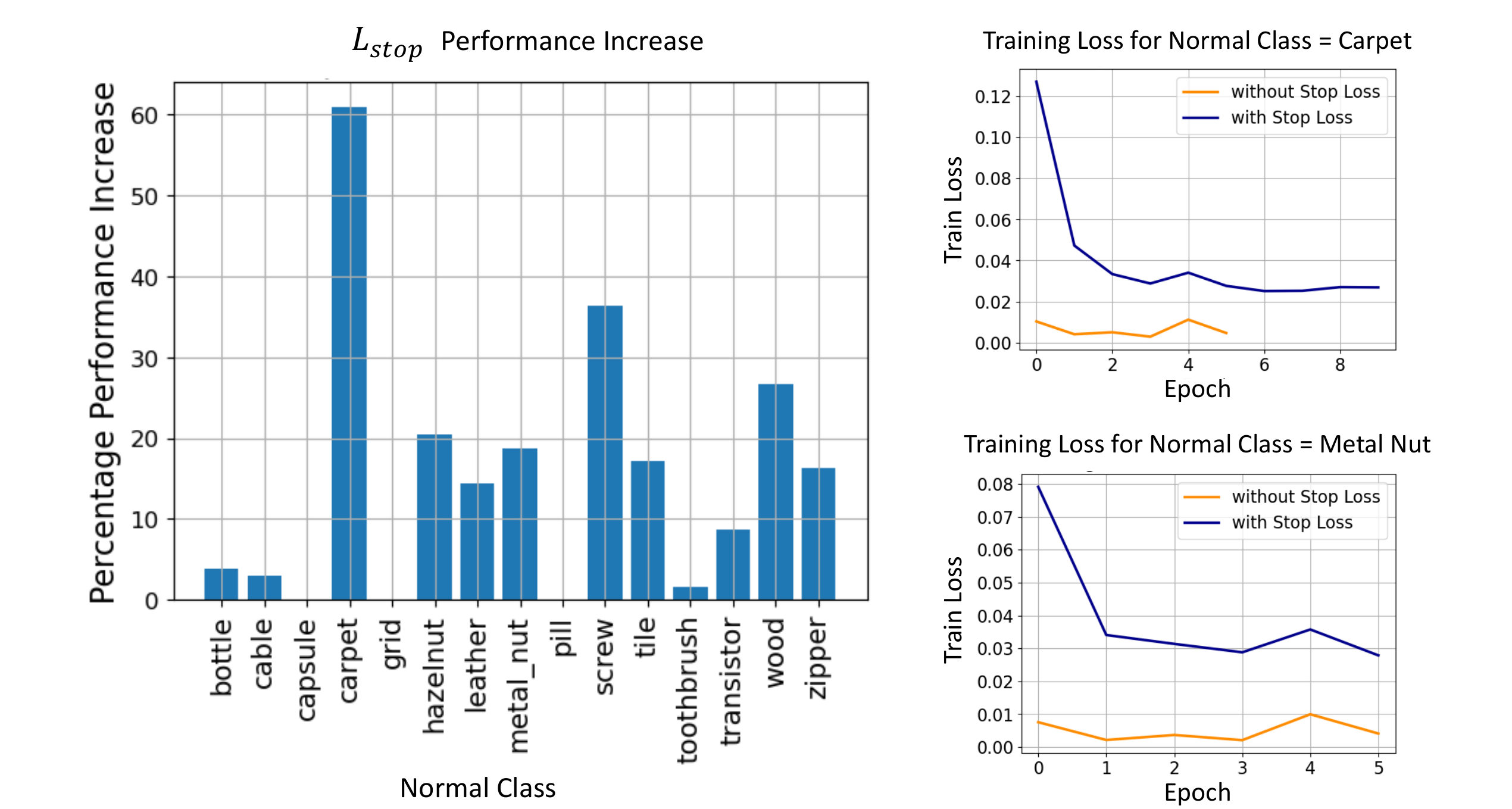}
  \caption{The barplot shows the percentage performance increase in average AUC over ten seeds as a result of including $L_{stop}$ in the loss function. The figure also depicts two examples of the training loss at each epoch of training with and without $L_{stop}$.}

\label{fig:ablation}
\end{figure}

A barplot in Figure \ref{fig:ablation} shows the percentage performance increase in average AUC over ten seeds for each normal class in the MVTec dataset as a result of including $L_{stop}$ in the loss function. The plot shows significant increases of up to 60\% in one case, demonstrating the importance of including $L_{stop}$ in the loss function. Representational collapse can be identified when the model immediately reaches an almost zero training loss \cite{simsiam}. The training plots in Figure \ref{fig:ablation} show that the train loss immediately achieves almost zero values without $L_{stop}$. However, the inclusion of $L_{stop}$ regularises the model to prevent representational collapse and boosts the model performance.

\subsection{Sensitivity Analysis on Reference Set, N and choice of Anchor}
\label{sensitivity}
\begin{figure}

  \includegraphics[width=\linewidth]{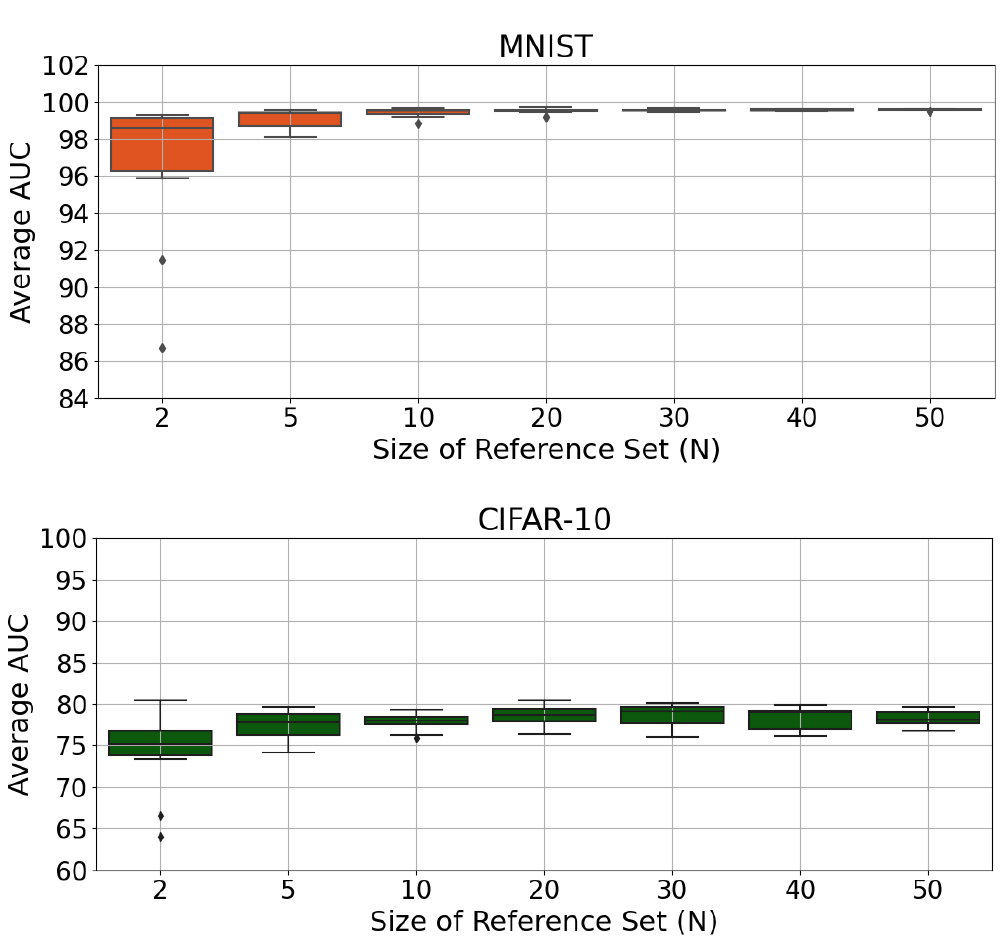}
  \caption{AUC in \% averaged over ten different randomly sampled Reference Sets for different values of N and different selections of the anchor.}

  \label{fig:ref}
\end{figure}

A sensitivity analysis was conducted to assess how the model performance is impacted depending on \textbf{(a)} the choice of Reference set, \textbf{(b)} choice of N and \textbf{(c)} choice of the training sample, $r_{a}$ that is used to compute the anchor. For each value of $N$, ten models were trained based on different randomly sampled Reference sets and randomly sampled $r_{a}$. Figure \ref{fig:ref} shows the model's performance on MNIST with the normal class set to zero and CIFAR-10 with the normal class set to \lq{}airplane\rq{}. The results show that the performance increases as $N$ is increased from two to five before plateauing at $N=5$ for both datasets. From this, we outline the following guidelines for choosing N. 
\begin{enumerate}

\item Increase N until the performance begins to plateau and there is little variability in model performance between choice of Reference Set. 
\item In the absence of labelled data, our experiments have shown $N=30$ to achieve good performance across all datasets.
\end{enumerate}

As there is little variability in model performance once $N$ is large enough (i.e. greater than $N=5$), we can also conclude that the selection of the Reference set and $r_{a}$ does not impact performance. Therefore, random sampling is an effective method for selecting the Reference set and $r_{a}$.

\section{Conclusion}

In this paper, we presented FewSOME, a new SOTA in the field of FSAD on normal data samples. By reporting on AUC, F1 scores, balanced accuracy and training times, we have demonstrated FewSOME's ability to perform at SOTA level having trained on a fraction of the data that existing approaches require. Our extensive experiments have shown that our proposed loss, \lq{}Stop Loss\rq{} prevents representational collapse and boosts performance. We have also demonstrated that FewSOME is robust to contaminated datasets and it is insensitive to choice of Reference Set and anchor.

\section*{Acknowledgements}
This work was funded by Science Foundation Ireland through the SFI Centre for Research Training in Machine Learning (Grant No. 18/CRT/6183). This work is supported by the Insight Centre for Data Analytics under Grant Number SFI/12/RC/2289 P2.

{\small
\bibliographystyle{ieee_fullname}
\bibliography{egbib}
}

\end{document}